\documentclass[10pt,twocolumn,letterpaper]{article}

\usepackage{cvpr}              % To produce the CAMERA-READY version
\definecolor{cvprblue}{rgb}{0.21,0.49,0.74}
\usepackage[pagebackref,breaklinks,colorlinks,allcolors=cvprblue]{hyperref}
\usepackage[most]{tcolorbox}
\tcbuselibrary{listings,breakable}

\newtcblisting{promptbox}[2][]{
breakable,
enhanced,
colback=gray!4,
colframe=black,
arc=2mm,
boxrule=0.8pt,
left=2mm,
right=2mm,
top=1mm,
bottom=1mm,
title=\textbf{#2},
fonttitle=\bfseries,
listing only,
listing options={
basicstyle=\ttfamily\footnotesize,
breaklines=true,
columns=fullflexible,
keepspaces=true,
showstringspaces=false
},
#1
}

\def\confName{CVPR}
\def\confYear{2026}

\title{ParseFixer: An Agentic Framework for Document Parsing via Selective Multimodal Correction}

\author{
LeKai Yu$^1$ \quad
Hao Liu$^1$ \quad
Kun Wang$^1$ \quad
Zhiran Li$^1$ \quad
Ruping Cao$^1$ \quad
Fan Liu$^2$ \quad
Yupeng Hu$^1$ \\
\\
$^1$Shandong University
$^2$Southeast University \\
{\tt\small \{kyleyue70, liuh90210, khylon.kun.wang, zhiranli325, caoruping657, liufancs\}@gmail.com} \\
{\tt\small huyupeng@sdu.edu.cn}}

\begin{document}
\maketitle
\begin{abstract} 
In this report, we present our third-place solution for the DataMFM Challenge Track 1: Document Parsing. This track requires models to recover structured Markdown documents from document page images while preserving textual content and document structure. To address the complementary requirements of accurate content recovery and faithful structure reconstruction, we propose \textbf{ParseFixer}, an agentic framework for backbone parsing and selective correction. ParseFixer consists of two key modules: Full-Page Backbone Parsing (FBP) and Agentic Selective Correction (ASC). FBP produces stable initial Markdown outputs with MinerU2.5 Pro, while ASC detects high-value parsing failures and repairs them through a verify-and-rollback correction process. By placing selective multimodal correction after open-source backbone parsing, ParseFixer improves the recovery of key document elements without rewriting reliable backbone predictions. On the test set, our final system achieves an overall score of 61.78 and ranks third in Track 1, demonstrating its effectiveness for accurate document parsing. Our code will be released at: \url{https://github.com/iLearn-Lab/CVPRW26-ParseFixer}. 
\end{abstract}    
\section{Introduction}

With the rapid growth of visually rich documents, document parsing has become a key step for converting page images into machine-readable representations~\cite{ouyang2025omnidocbench, dong2026doc, kondic2026chartnet, yu2026omniparser, liu2026chartlens}. Recent advances in multimodal understanding have improved the ability of models to interpret visual content with textual semantics~\cite{wang2024explicit, liu2025gaming, hu2026glance, wang2025redundancy, li2025mist, hu2026visual, xiang2025dkdm, xiang2026tina}. However, document parsing goes beyond general visual-textual understanding. It requires models to recover page content while preserving the structural organization behind it. This makes the task inherently layout-sensitive. A document page may involve dense regions, irregular layouts, or format-sensitive components, where local recognition errors can easily affect the recovery of content hierarchy and reading order. Therefore, document parsing has become an important problem in multimodal document intelligence, as structured document representations can further benefit general multimodal learning scenarios, from vision-language pretraining to video-language understanding~\cite{liu2025curmim, hu2021video, hu2021coarse, zhao2023cogcn, li2026unim, hu2023semantic, li2025dcount, wang2026cross}.

\begin{figure}[t]
\centering
\includegraphics[width=0.46\textwidth]{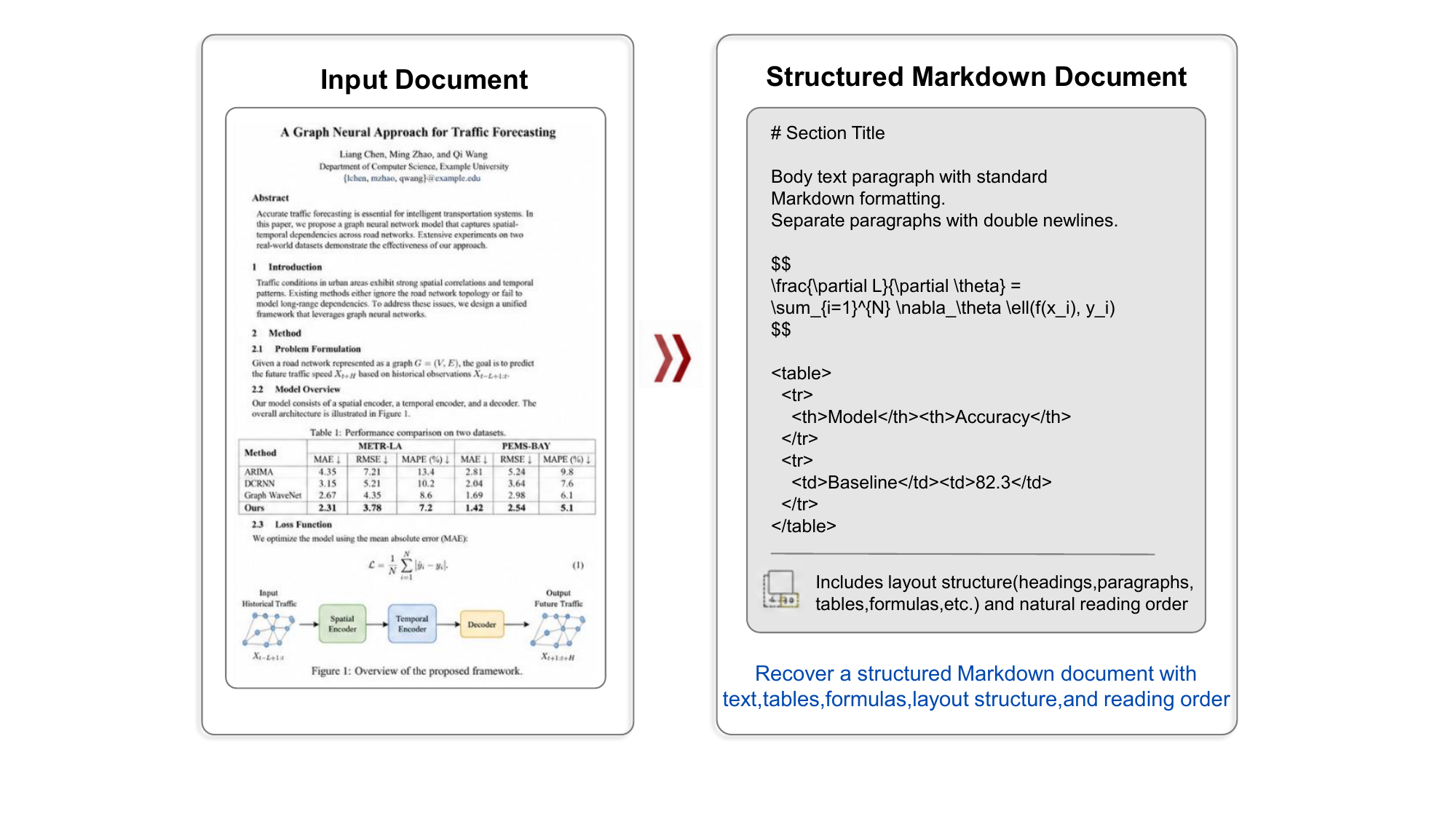}
\vspace{-2mm}
\caption{Task definition of DataMFM Challenge Track 1. Given document page images, the model is required to recover document-level Markdown files that preserve page content and structural order.}
\vspace{-6mm}
\label{fig:task_definition}
\end{figure}

The DataMFM Challenge Track 1\footnote{https://datamfm.github.io/challenge.html} focuses on document parsing in multimodal document scenarios. As shown in Figure~\ref{fig:task_definition}, given document page images, participating models are required to produce document-level Markdown files. The output should faithfully preserve the page content and its structural order. Therefore, this track is not only about recognizing what appears on the page, but also about reconstructing how the page is organized.

Despite the progress of document parsing models and multimodal foundation models~\cite{cui2025paddleocr, cui2026paddleocr, li2023blip, duan2026glm, li2022blip}, directly using a single model to generate the final Markdown remains unreliable for this challenge. The core limitation is that document parsing requires both content recognition and structure reconstruction. A model may correctly recognize local text but still fail to place it in the right document context. Conversely, a stronger multimodal model may handle difficult regions better, but full-page regeneration can introduce rewriting, summarization, or hallucinated content. This evaluation setting exposes two practical challenges:
1) \textbf{Structure-Preserving Parsing.} The parsed result should preserve the logical organization of the original page rather than only transcribe local text. In other words, local recognition is not equivalent to document-level reconstruction.
2) \textbf{Controlled Failure Correction.} Most pages can already be parsed reasonably well by a strong backbone parser, while only a small portion requires additional correction. In other words, difficult regions should be repaired selectively instead of rewriting the entire document.

\begin{figure}[t]
\centering
\includegraphics[width=0.46\textwidth]{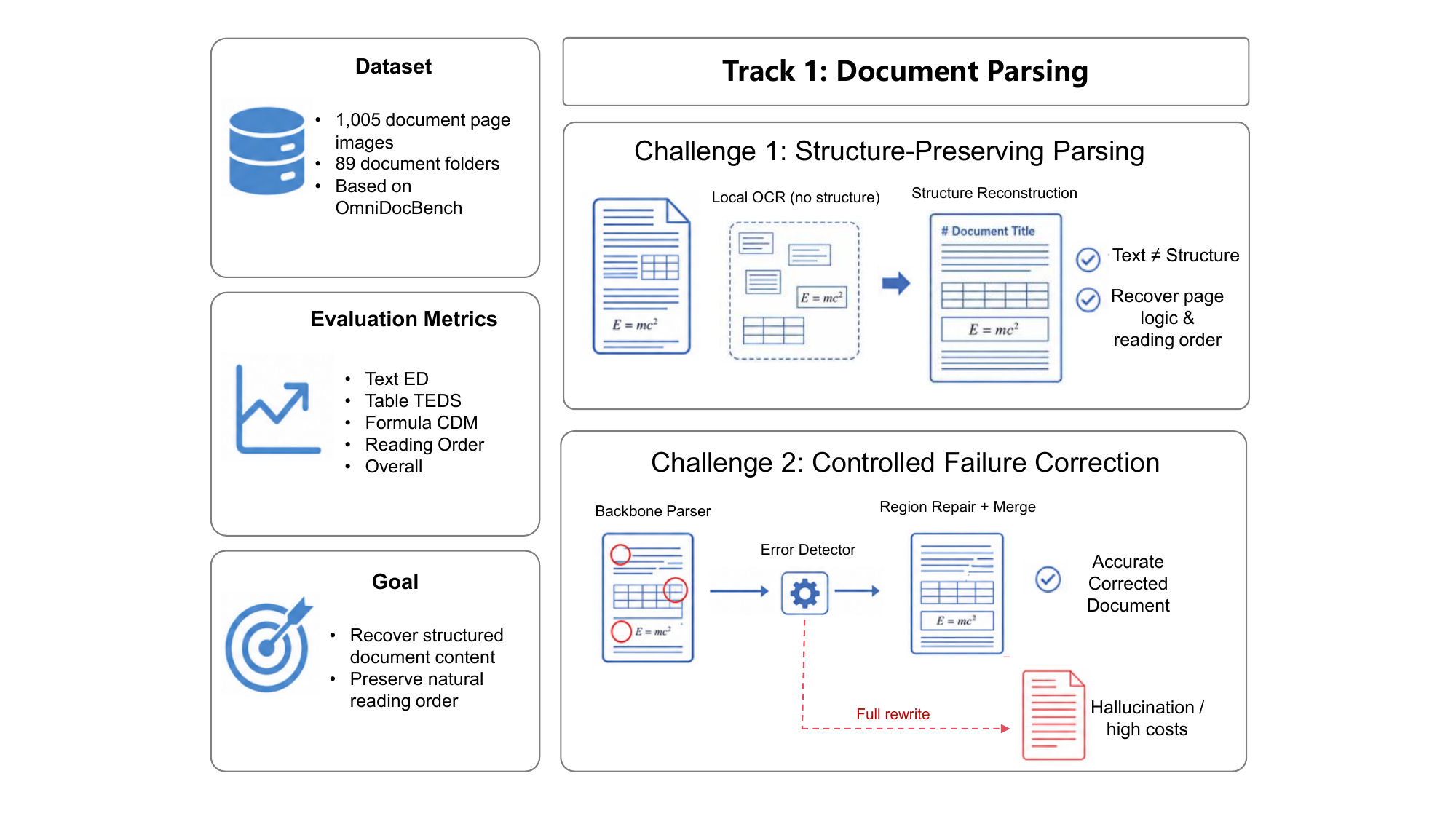}
\vspace{-2mm}
\caption{Motivation of our solution for DataMFM Track 1.}
\vspace{-6mm}
\label{fig:motivation}
\end{figure}

To tackle these challenges, we adopt a selective correction strategy for document parsing. The core intuition is that the initial parsing result should not be entirely discarded when only part of it is unreliable. Instead, reliable predictions should be kept, while suspicious outputs should be corrected under explicit verification. To this end, we propose \textbf{ParseFixer}, an agentic framework for document parsing with selective multimodal correction. ParseFixer consists of two key modules: Full-Page Backbone Parsing (FBP) and Agentic Selective Correction (ASC). FBP generates the initial Markdown output with MinerU2.5 Pro~\cite{wang2026mineru2}, while ASC detects unreliable parsing results and corrects them through a verify-and-rollback process. By placing controlled correction after backbone parsing, ParseFixer improves final Markdown quality without unnecessarily rewriting reliable predictions.

In summary, our main contributions are threefold:
\begin{itemize}[leftmargin=*]
\item We propose \textbf{ParseFixer}, an agentic framework for document parsing that treats Markdown generation as a backbone parsing followed by selective correction process.

\item We introduce a verify-and-rollback correction strategy to preserve reliable parsing results while revising unreliable pages or regions only when necessary.

\item We validate the proposed framework on the DataMFM Challenge Track 1, where our final model ranks \textit{third} among submitted solutions.
\end{itemize}

\section{Method}

\begin{figure*}[t]
\centering
\includegraphics[width=0.7\textwidth]{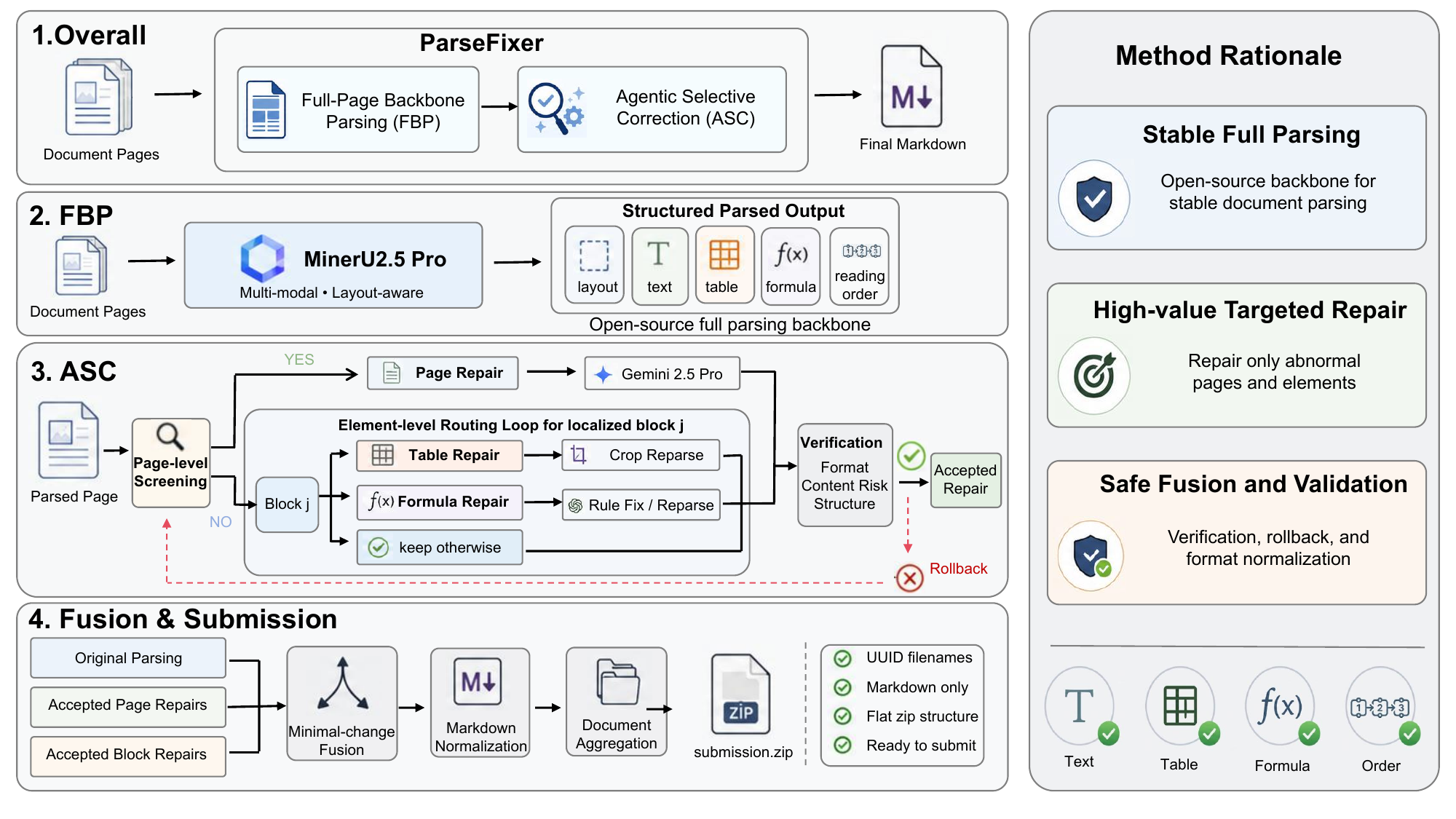}
\vspace{-2mm}
\caption{Overview of ParseFixer. Given document page images, ParseFixer first uses FBP to generate initial Markdown outputs and structured parsing information with MinerU2.5 Pro. Then, ASC detects unreliable pages or local elements and performs selective correction through multimodal re-parsing, crop-level correction, rule-based repair, and candidate verification. The final document-level Markdown files are obtained by accepting verified corrections and preserving reliable backbone predictions.}
\vspace{-4mm}
\label{fig:method_overview}
\end{figure*}

In this section, we first formulate the document parsing task and present the overall architecture of ParseFixer, as illustrated in Figure~\ref{fig:method_overview}. We then introduce the two key modules of ParseFixer, namely Full-Page Backbone Parsing (FBP) and Agentic Selective Correction (ASC). Finally, we summarize the inference pipeline used for the final submission.

\subsection{Task Formulation}

Given a document $\mathcal{D}$ with $N$ page images, DataMFM Track 1 requires the model to produce a document-level Markdown file. We denote the input document as:
\begin{equation}
    \mathcal{D}=\{I_1,I_2,\ldots,I_N\},
\end{equation}
where $I_i$ denotes the $i$-th page image. The prediction target is a sequence of page-level Markdown outputs:
\begin{equation}
    Y=\{Y_1,Y_2,\ldots,Y_N\},
\end{equation}
where $Y_i$ denotes the Markdown result of page $I_i$. The final document-level Markdown is obtained by merging page-level outputs according to the original page order.

\subsection{Full-Page Backbone Parsing}

As shown in Figure~\ref{fig:method_overview}, FBP serves as the initial parsing stage of ParseFixer. Given a page image $I_i$, we use MinerU2.5 Pro~\cite{wang2026mineru2} as the backbone parser to produce the page-level Markdown output and layout-aware parsing information:
\begin{equation}
(Y_i^0,Z_i^0)=\mathrm{FBP}(I_i),
\end{equation}
where $Y_i^0$ denotes the initial Markdown output of page $I_i$, and $Z_i^0$ denotes the structured parsing information used for subsequent selective correction. For the whole document, the initial prediction is:
\begin{equation}
Y^0=\{Y_1^0,Y_2^0,\ldots,Y_N^0\}.
\end{equation}

FBP performs full-page parsing to recover both visual content and document organization. The resulting layout-aware output guides Markdown generation, so that headings, tables, and formulas can be represented with appropriate structural formats. To make this information accessible for subsequent correction, we represent the structured parsing information as:
\begin{equation}
Z_i^0=\{E_i^0,B_i^0,O_i^0\},
\end{equation}
where $E_i^0=\{e_{i,j}^0\}_{j=1}^{M_i}$ denotes the parsed document elements on page $I_i$, $B_i^0=\{b_{i,j}^0\}_{j=1}^{M_i}$ denotes their corresponding bounding boxes, $O_i^0$ denotes the predicted reading order, and $M_i$ is the number of parsed layout blocks on page $I_i$. The initial Markdown output is generated by arranging these elements according to the predicted order:
\begin{equation}
Y_i^0=\Gamma(E_i^0,B_i^0,O_i^0),
\end{equation}
where $\Gamma(\cdot)$ denotes the layout-aware Markdown conversion process, which orders parsed blocks by reading order and formats them according to their layout types.

We keep $Y^0$ as the default prediction because the backbone parser provides stable full-page parsing for most pages. Meanwhile, $Z_i^0$ provides localization cues for correcting format-sensitive regions in ASC. Therefore, FBP provides the parsing foundation of ParseFixer, while ASC is only invoked for unreliable pages or local elements.

\subsection{Agentic Selective Correction}

As shown in Figure~\ref{fig:method_overview}, ASC is introduced after FBP to revise unreliable parsing results while preserving reliable backbone predictions. Since most pages can already be parsed reasonably well by the backbone parser, ASC does not perform correction by default. Instead, it first diagnoses explicit parsing failures and then applies correction only to the affected page or local element.

Given the page image $I_i$, the initial Markdown output $Y_i^0$, and the layout-aware parsing information $Z_i^0$, ASC performs explicit failure diagnosis based on \textit{page-level quality checks} and \textit{element-level format checks}. The former identifies holistic parsing failures, while the latter focuses on format-sensitive local regions, especially tables and formulas. ASC casts failure diagnosis as a set of binary rule-based trigger conditions, which determine whether correction is necessary.

The page-level routing action is formulated as:
\begin{equation}
a_i^{p} =
\begin{cases}
\phi_{\text{page}}, & \mathcal{C}_i^{p}=1, \\
\texttt{Proceed}, & \mathcal{C}_i^{p}=0,
\end{cases}
\end{equation}
where $\mathcal{C}_i^{p} \in \{0,1\}$ denotes the page-level correction trigger. It is activated when the page output exhibits severe parsing failures, such as structural truncation, abnormal empty output, or globally disordered Markdown generation. If $\mathcal{C}_i^{p}=1$, ASC performs page-level correction directly. Otherwise, it proceeds to the element-level routing loop.

For each localized layout block $j$ on page $I_i$, the element-level routing action is formulated as:
\begin{equation}
a_{i,j} =
\begin{cases}
\phi_{\text{table}}, & \mathcal{C}_{i,j}^{t}=1, \\
\phi_{\text{formula}}, & \mathcal{C}_{i,j}^{t}=0 \land \mathcal{C}_{i,j}^{f}=1, \\
\phi_{\text{keep}}, & \mathcal{C}_{i,j}^{t}=0 \land \mathcal{C}_{i,j}^{f}=0,
\end{cases}
\end{equation}
where $\mathcal{C}_{i,j}^{t}\in\{0,1\}$ and $\mathcal{C}_{i,j}^{f}\in\{0,1\}$ denote the table-level and formula-level correction triggers of block $j$, respectively. The table trigger is activated by malformed HTML table structures, missing cells, structural truncation, or table-like text patterns that satisfy the pseudo-table upgrading rules. The formula trigger is activated by unmatched formula delimiters, unbalanced braces, mismatched LaTeX environments, abnormal LaTeX commands, or formula-like text patterns that satisfy the pseudo-formula upgrading rules. When both table and formula triggers are activated for the same block, the table branch is prioritized because table-level structural recovery may also preserve formula content inside the table. The functions $\phi_{\text{page}}$, $\phi_{\text{table}}$, and $\phi_{\text{formula}}$ denote the targeted correction operators, while $\phi_{\text{keep}}$ indicates that the original block is preserved.

For page-level correction, ASC uses multimodal re-parsing under strict Markdown constraints:
\begin{equation}
\hat{Y}_i^{p} = \phi_{\text{page}}(I_i, Y_i^0, P_{\text{page}}),
\end{equation}
where $P_{\text{page}}$ denotes the page-level correction prompt provided in the supplementary material. This branch is used when the initial Markdown $Y_i^0$ is unreliable at the page level. Under this strategy, the original full-page image $I_i$ is fed into Gemini 2.5 Pro~\cite{comanici2025gemini} to perform visual OCR and structural reconstruction. The initial output $Y_i^0$ serves only as an auxiliary reference for the failure mode. The prompt requires the model to recover visible page content and strictly avoid explanations, summaries, translations, or unsupported additions.

The page-level candidate is verified before being accepted:
\begin{equation}
v_i^{p}=\mathrm{Verify}_{\text{rule}}(I_i,Y_i^0,\hat{Y}_i^{p};\phi_{\text{page}}),
\end{equation}
where $v_i^{p}\in\{0,1\}$ indicates whether the page-level candidate is accepted. The output of the page-level branch is:
\begin{equation}
\bar{Y}_i^{p}=
\begin{cases}
\hat{Y}_i^{p}, & v_i^{p}=1,\\
Y_i^0, & v_i^{p}=0.
\end{cases}
\end{equation}

If page-level correction is not triggered, ASC performs element-level correction. The element-level process is initialized as:
\begin{equation}
Y_i^{(0)}=Y_i^0.
\end{equation}
ASC then scans localized blocks according to the predicted reading order $O_i^0$.

For table correction, ASC first locates the suspicious table region according to the bounding box $b_{i,j}^0$ in $Z_i^0$. Let
\begin{equation}
I_{i,j}^{t}=\mathrm{Crop}(I_i,b_{i,j}^0)
\end{equation}
denote the cropped table region of block $j$. The corrected table candidate is generated as:
\begin{equation}
\hat{T}_{i,j} = \phi_{\text{table}}(I_{i,j}^{t}, T_{i,j}^0, P_{\text{table}}),
\end{equation}
where $T_{i,j}^0$ denotes the initial table-related output of block $j$, and $P_{\text{table}}$ denotes the table correction prompt detailed in the supplementary material. ASC first uses the localized cropping tool of MinerU2.5 Pro for stable structural extraction. If localized parsing fails to resolve the structural anomaly, the module falls back to Gemini 2.5 Pro with strict instructions to output a valid HTML table. If a plain text block is diagnosed as a potential borderless table, it is promoted to a pseudo-table element and sent to the same localized table repair pipeline.

For formula correction, ASC first applies deterministic LaTeX repair:
\begin{equation}
\tilde{F}_{i,j}=\rho_{\text{latex}}(F_{i,j}^0),
\end{equation}
where $F_{i,j}^0$ is the initial formula-related output of block $j$, and $\rho_{\text{latex}}(\cdot)$ denotes the rule-based LaTeX repair function. This step fixes minor formatting issues without invoking large models, such as delimiter normalization, LaTeX command correction, bracket balancing, and currency-symbol escaping.

The formula candidate is then determined as:
\begin{equation}
\bar{F}_{i,j}=
\begin{cases}
\tilde{F}_{i,j}, & \mathrm{Valid}_{\text{latex}}(\tilde{F}_{i,j})=1,\\
\phi_{\text{formula}}(I_{i,j}^{f},\tilde{F}_{i,j},P_{\text{formula}}), & \mathrm{Valid}_{\text{latex}}(\tilde{F}_{i,j})=0,
\end{cases}
\end{equation}
where
\begin{equation}
I_{i,j}^{f}=\mathrm{Crop}(I_i,b_{i,j}^0)
\end{equation}
denotes the cropped formula region, and $P_{\text{formula}}$ denotes the LaTeX-only correction prompt detailed in the supplementary material. When deterministic repair is insufficient, the candidate formula is first generated by the local crop operator of MinerU2.5 Pro. If syntax invalidity persists, ASC falls back to GPT-5.5~\cite{openai2026gpt55systemcard} under instructions to output raw LaTeX only, without contextual descriptions or explanations. Text blocks promoted by the pseudo-formula mechanism follow the same localized re-parsing workflow.

For element-level correction, the corrected local candidate is inserted back into the current page-level Markdown output to form a page-level candidate:
\begin{equation}
\hat{Y}_{i,j}=
\begin{cases}
\mathrm{Insert}(Y_i^{(j-1)},j,\hat{T}_{i,j}), & a_{i,j}=\phi_{\text{table}},\\
\mathrm{Insert}(Y_i^{(j-1)},j,\bar{F}_{i,j}), & a_{i,j}=\phi_{\text{formula}},\\
Y_i^{(j-1)}, & a_{i,j}=\phi_{\text{keep}}.
\end{cases}
\end{equation}
Here, $\mathrm{Insert}(\cdot)$ replaces only the selected local block while preserving the remaining page content and the predicted reading order.

ASC does not directly accept a corrected candidate. For non-keep actions, the candidate is checked by a rule-based verification procedure:
\begin{equation}
v_{i,j}=\mathrm{Verify}_{\text{rule}}(I_i,Y_i^{(j-1)},\hat{Y}_{i,j};a_{i,j}),
\end{equation}
where $v_{i,j}\in\{0,1\}$ indicates whether the local correction is accepted. The iterative element-level update is:
\begin{equation}
Y_i^{(j)}=
\begin{cases}
\hat{Y}_{i,j}, & a_{i,j}\neq\phi_{\text{keep}} \land v_{i,j}=1,\\
Y_i^{(j-1)}, & \text{otherwise}.
\end{cases}
\end{equation}

The verification step is adapted to different correction types. For page-level candidates, ASC rejects outputs containing non-source descriptions, explanations, or unsupported content. For table candidates, ASC checks whether the result contains a valid HTML table with legal rows and cells, ensuring that it is neither an empty structure nor a pipe-format table. For formula candidates, ASC verifies LaTeX-only formatting and the consistency of delimiters, braces, and environments. In addition to format verification, ASC compares the candidate with the previous page output. A candidate is rejected if it causes obvious content loss, disrupts reading order, or introduces a large amount of non-source text. Therefore, the verification process can be viewed as a constrained selection between the current output and the corrected candidate, where format validity, content faithfulness, and structural preservation are jointly considered.

The final page-level output of ASC is:
\begin{equation}
Y_i^\ast=
\begin{cases}
\bar{Y}_i^{p}, & \mathcal{C}_i^{p}=1,\\
Y_i^{(M_i)}, & \mathcal{C}_i^{p}=0.
\end{cases}
\end{equation}
In this way, ASC improves unreliable pages or local regions while preserving reliable backbone predictions.

\subsection{Overall Inference}

During inference, ParseFixer follows the staged process illustrated in Figure~\ref{fig:method_overview}. For each document page image $I_i$, FBP first generates the initial Markdown output and layout-aware parsing information:
\begin{equation}
(Y_i^0,Z_i^0)=\mathrm{FBP}(I_i).
\end{equation}
ASC then performs explicit failure diagnosis based on page-level quality checks and element-level format checks.

If the page-level trigger is activated, ASC applies page-level correction, verifies the corrected candidate, and either accepts the verified candidate or falls back to the original backbone output. If the page-level trigger is not activated, ASC iterates over localized layout blocks. For each block, it decides whether to keep the original output, repair a table, or repair a formula. Each corrected local candidate is inserted back into the page-level Markdown and accepted only after rule-based verification. If no unreliable page or local element is detected, the initial Markdown output is directly preserved.

After all pages are processed, the accepted page-level Markdown outputs are normalized with deterministic formatting rules. The normalization step only removes formatting noise and does not rewrite semantic content. Finally, the page-level results are merged according to the original page order:
\begin{equation}
Y^\ast=\{Y_1^\ast,Y_2^\ast,\ldots,Y_N^\ast\}.
\end{equation}
The resulting document-level Markdown files are packed into the official submission format.
\section{Experiments}

\subsection{Experimental Settings}

\textbf{Dataset.} We evaluate ParseFixer on the official DataMFM Challenge Track 1 document parsing dataset. The dataset is newly prepared based on OmniDocBench~\cite{ouyang2025omnidocbench} and contains 1,005 page images organized under 89 document folders. It supports page-level parsing evaluation over natural text, tables, formulas, layout structure, and reading order.

\textbf{Evaluation Metrics.}
Following the official evaluation protocol of DataMFM Challenge Track 1, we evaluate each submission using four metrics: Text ED, Table TEDS, Formula CDM, and Reading Order. Text ED measures text-level edit distance and lower values indicate better performance. Table TEDS, Formula CDM, and Reading Order evaluate table structure recovery, formula recognition, and document ordering quality, respectively. The final Overall score is computed by the official evaluation script.

\textbf{Implementation Details.}
We use MinerU2.5 Pro~\cite{wang2026mineru2} as the full-page backbone parser in FBP to generate the initial Markdown output and layout-aware parsing information. In ASC, selective correction is triggered only for unreliable pages or local blocks detected by page-level quality checks and element-level format checks. Gemini 2.5 Pro~\cite{comanici2025gemini} is used for page-level multimodal re-parsing and as a fallback model for table correction. MinerU2.5 Pro is also used for localized crop-level re-parsing of table and formula regions. GPT-5.5~\cite{openai2026gpt55systemcard} is used only as a fallback model for formula correction. All candidate corrections are verified before being accepted; otherwise, ASC rolls back to the original MinerU2.5 Pro output.

\subsection{Leaderboard Comparison}

Table~\ref{tab:leaderboard} reports the official leaderboard comparison of DataMFM Challenge Track 1. Our team, zed, ranks \textit{third} among all submitted solutions with an Overall score of 61.78. Compared with the fourth-ranked team, our solution improves the Overall score by 1.29 points. In terms of sub-metrics, our system achieves the best Text ED and Reading Order among all teams, indicating its advantage in text recovery and document structure preservation.

\begin{table}[t]
\centering
\caption{Leaderboard on DataMFM Challenge Track 1, sorted by Overall score in ascending order. Text ED is better when lower, while the other metrics are better when higher.}
\vspace{-2mm}
\label{tab:leaderboard}
\resizebox{0.99\linewidth}{!}{%
\begin{tabular}{lccccc}
\toprule
Team & Text ED $\downarrow$ & Table TEDS $\uparrow$ & Formula CDM $\uparrow$ & Reading Order $\uparrow$ & Overall $\uparrow$ \\
\midrule
SFD & 0.37 & 12.19 & 0.00 & 45.61 & 30.27 \\
hskl18 & 0.37 & 22.13 & 0.52 & 51.38 & 34.14 \\
HHHHHHHHHH & 0.27 & 65.47 & 0.65 & 53.80 & 48.26 \\
Wind\_Rain\_Tower & 0.15 & 59.56 & 0.58 & 61.59 & 51.62 \\
sig & 0.18 & 84.99 & 0.41 & 63.31 & 57.74 \\
ytttttt & 0.16 & \textbf{91.05} & 0.62 & 60.74 & 59.01 \\
anmspro & 0.18 & 82.03 & 0.51 & 72.62 & 59.34 \\
cdefg & 0.16 & 80.77 & 2.71 & 70.91 & 59.64 \\
dennis & 0.13 & 81.82 & 0.41 & 72.92 & 60.49 \\
\textbf{zed (Ours)} & \textbf{0.10} & 80.55 & 0.94 & \textbf{75.48} & 61.78 \\
durgasandeep & 0.17 & 85.20 & 15.73 & 64.74 & 62.08 \\
Zhiheng & 0.16 & 82.42 & \textbf{19.30} & 75.44 & \textbf{65.37} \\
\bottomrule
\end{tabular}%
}
\vspace{-2mm}
\end{table}

\subsection{Candidate Parser Comparison}

We first compare several candidate parsers before building the final model. This experiment is used to select a stable full-page parsing backbone rather than treating model selection as the main contribution. As shown in Table~\ref{tab:candidate_parser}, MinerU2.5 Pro achieves the best Overall score among the open-source parsers, with clear advantages in Table TEDS and Reading Order. Although the closed-source multimodal parser obtains competitive Text ED and Table TEDS, its Overall score is close to MinerU2.5 Pro and it does not provide a stable improvement under full-page generation. Therefore, we select MinerU2.5 Pro as the FBP backbone and use closed-source multimodal models only for selective correction in ASC.

\begin{table}[t]
\centering
\caption{Candidate parser comparison. Text ED is better when lower, while the other metrics are better when higher.}
\vspace{-2mm}
\label{tab:candidate_parser}
\resizebox{\linewidth}{!}{%
\begin{tabular}{lccccc}
\toprule
Parser & Text ED $\downarrow$ & Table TEDS $\uparrow$ & Formula CDM $\uparrow$ & Reading Order $\uparrow$ & Overall $\uparrow$ \\
\midrule
PaddleOCR-VL~\cite{cui2026paddleocr} & 0.1939 & 70.59 & 0.25 & 62.30 & 53.44 \\
GPT-5.5~\cite{openai2026gpt55systemcard} & \textbf{0.1494} & \textbf{83.80} & 0.56 & 61.87 & 57.82 \\
\textbf{MinerU2.5 Pro}~\cite{wang2026mineru2} & 0.1641 & 80.55 & 0.41 & \textbf{67.36} & \textbf{57.98} \\
\bottomrule
\end{tabular}}
\vspace{-2mm}
\end{table}

\subsection{Component Analysis}

Since this track evaluates the final submitted Markdown files rather than a fixed trainable model, our experiments are structured as a competition-style progressive refinement analysis instead of strictly independent ablation studies. As shown in Table~\ref{tab:component_analysis}, each setting corresponds to a submission candidate obtained by adding a specific correction strategy on top of the previous configuration.

Starting from the MinerU baseline (S0), basic layout normalizations such as heading and formula formatting (S1, S2) provide initial stabilization for text recovery and reading order. Subsequent integration of the OCR degeneration repair module (S3) specifically targets repetitive and garbled texts, steadily reducing the Text ED metric. Additionally, we introduce a pseudo-formula promotion mechanism (S4) to recover missed mathematical expressions, providing a marginal but positive contribution to the overall score.

For broader page-level issues, the anomaly routing module (S5, S6) effectively mitigates parsing failures on complex pages. S7 represents the best intermediate candidate observed during the refinement process, while S8 corresponds to the official submitted version. Although S8 slightly decreases Table TEDS, it improves Text ED and Reading Order and was selected as the final submitted configuration under the actual submission constraints.

\begin{table}[t]
\centering
\caption{Component analysis of ParseFixer on DataMFM Challenge Track 1. Text ED is better when lower, while the other metrics are better when higher.}
\vspace{-2mm}
\label{tab:component_analysis}
\resizebox{\linewidth}{!}{%
\begin{tabular}{lccccc}
\toprule
Setting & Text ED $\downarrow$ & Table TEDS $\uparrow$ & Formula CDM $\uparrow$ & Reading Order $\uparrow$ & Overall $\uparrow$ \\
\midrule
S0: MinerU original output & 0.1641 & 80.55 & 0.41 & 67.36 & 57.98 \\
S1: S0 + formula formatting \& heading repair & 0.1559 & 80.55 & 0.41 & 68.18 & 58.39 \\
S2: S1 + formula format repair & 0.1559 & 80.55 & 0.41 & 68.18 & 58.39 \\
S3: S1 + OCR degeneration repair & 0.1310 & 80.55 & 0.41 & 70.64 & 59.63 \\
S4: S3 + pseudo-formula promotion & 0.1300 & 80.66 & \textbf{0.94} & 70.67 & 59.82 \\
S5: S3 + empty/short page repair & 0.1028 & 80.55 & 0.41 & 74.77 & 61.36 \\
S6: S1 + page and formula repair & 0.0997 & 80.55 & \textbf{0.94} & 75.44 & 61.74 \\
S7: S3 + page, formula, and table repair & 0.0993 & \textbf{80.92} & \textbf{0.94} & 75.40 & \textbf{61.83} \\
S8: Final submission & \textbf{0.0986} & 80.55 & \textbf{0.94} & \textbf{75.48} & 61.78 \\
\bottomrule
\end{tabular}}
\vspace{-4mm}
\end{table}

\subsection{Qualitative Analysis}

\begin{figure}[t]
\centering
\includegraphics[width=0.40\textwidth]{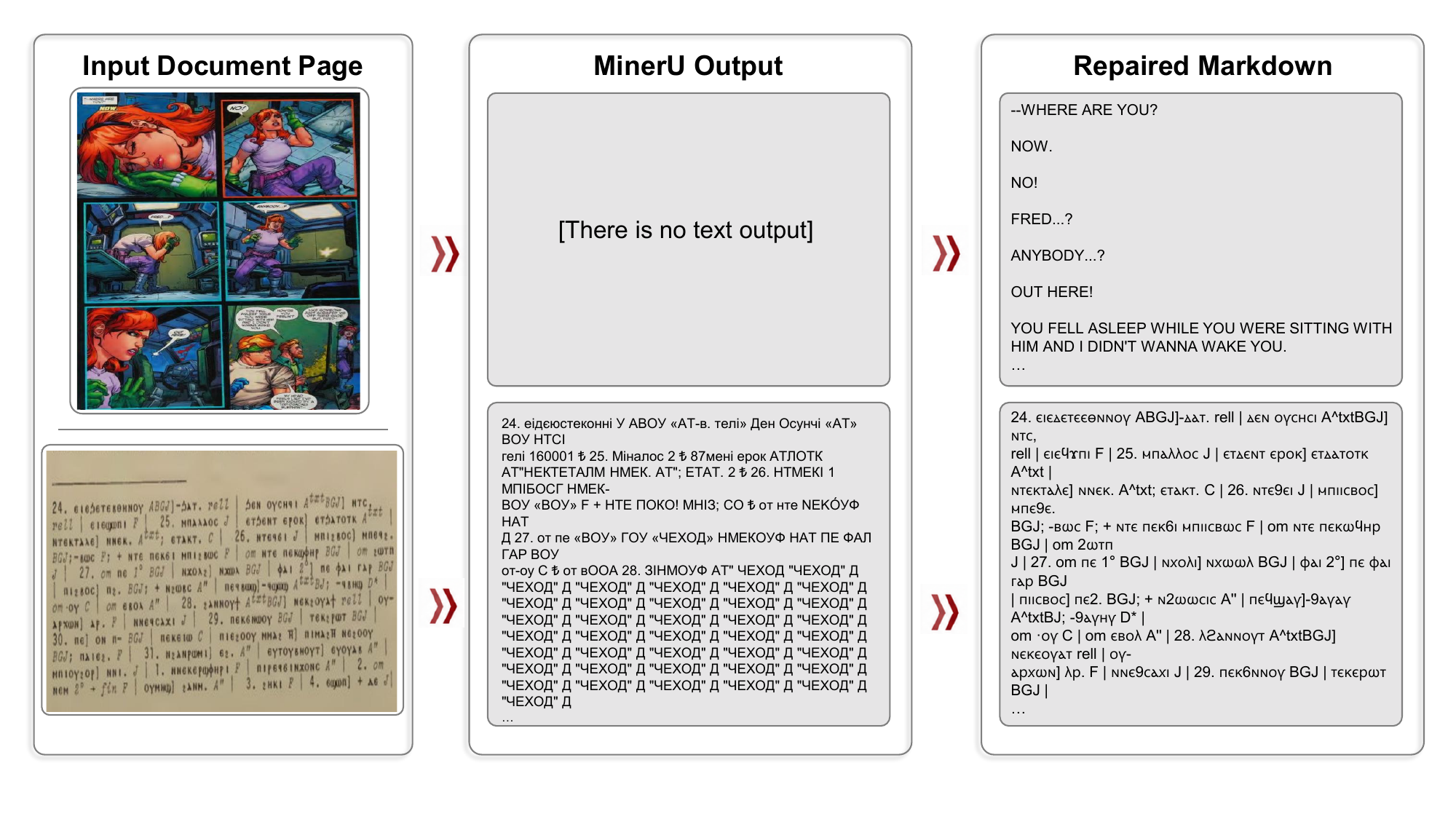}
\vspace{-4mm}
\caption{Qualitative examples of page-level failure modes addressed by ParseFixer.}
\vspace{-6mm}
\label{fig:case_1}
\end{figure}

We provide qualitative examples to further analyze representative page-level failure modes addressed by ParseFixer. As shown in Figure~\ref{fig:case_1}, although challenging pages differ significantly in their visual characteristics—such as manga-style pages, rare scripts, low-resolution scans, dense text, and complex layouts—they often exhibit similar degradation symptoms in the initial Markdown outputs. These symptoms include empty outputs, extremely short texts, heavy repetition, garbled text, miscoded symbols, and massive omissions of visible content. Therefore, the key to page-level recovery lies in identifying these observable degradation symptoms and performing constrained full-page re-parsing when necessary.

\textbf{Empty or extremely short output.}
For certain pages, although the backbone parser completes the inference process, the generated Markdown contains significantly less content than what is visibly present on the page, or even lacks any valid text. This typically implies that the model failed to parse the core page content. Such issues are highly prevalent in manga-style pages or low-resolution scans. ParseFixer flags these pages as high-risk page-level samples and invokes a multimodal large model to recover the Markdown content directly from the page image.

\textbf{Repetitive output.}
In some cases, the initial Markdown exhibits continuously repeating phrases, sentences, or paragraphs. This failure pattern typically indicates that the backbone parser has fallen into a degeneration loop during the text generation stage. While simple deduplication rules can remove some repetitive contents, they are prone to over-correcting and mistakenly deleting legitimate repetitive structures inherent to the original text. Therefore, ParseFixer treats severe repetition as a catastrophic page-level parsing failure and leverages full-page re-parsing to obtain a more stable candidate result.

\textbf{Garbled characters, miscoded symbols, and script confusion.}
For pages featuring rare scripts, cross-lingual texts, or poor image quality, the initial Markdown outputs may contain a high density of anomalous characters, incorrect scripts, or text that is glaringly inconsistent with the original page. These errors are remarkably difficult to fix reliably via post-processing rules since the degradation has already occurred during the visual recognition stage. ParseFixer thus treats garbled text and miscoded symbols as page-level anomaly signals, triggering multimodal re-parsing to accurately reconstruct the visible content of the page from scratch.

In summary, these qualitative examples demonstrate that while manga-style pages, rare scripts, low-resolution scans, dense text, and complex layouts stem from different underlying causes, they manifest as highly similar page-level output degradations. Unifying these difficult samples into a single page-level anomaly correction pipeline enables the system to robustly handle diverse full-page failures without introducing a complex array of hand-crafted rules.

\section{External Resource Disclosure}

Our final ParseFixer uses MinerU2.5 Pro~\cite{wang2026mineru2} as the full-page document parsing backbone. External multimodal models are used only within ASC for selective correction. Specifically, Gemini 2.5 Pro~\cite{comanici2025gemini} is used for page-level correction and fallback table correction, while GPT-5.5~\cite{openai2026gpt55systemcard} is used for fallback formula correction. No additional manually annotated document parsing labels are introduced beyond the released challenge resources.

\section{Conclusion}

In this report, we present ParseFixer, our third-place solution for DataMFM Challenge Track 1. The core insight of ParseFixer is that document parsing should not rely on full-page regeneration for all pages. Instead, reliable backbone predictions should be preserved, while unreliable pages or local elements should be corrected selectively under explicit verification. To this end, ParseFixer combines Full-Page Backbone Parsing with Agentic Selective Correction, using MinerU2.5 Pro as the parsing foundation and a verify-and-rollback mechanism to control multimodal correction. Our final system achieves an Overall score of 61.78 and ranks \textit{third} in Track 1.

{
    \small
    \bibliographystyle{ieeenat_fullname}
    \bibliography{main}
}

\clearpage
\setcounter{page}{1}
\maketitlesupplementary

% ---------- Add to the preamble ----------

\setcounter{section}{0}

\section{Prompt Details}
\label{app:prompts}

This supplementary material provides the prompt templates used in ParseFixer for DataMFM Challenge Track 1: Document Parsing.All prompts follow the same principle: output Markdown or the required element format only, preserve visible document content, avoid explanations or summaries, and reject unsupported additions through verification and rollback.

\subsection{GPT5.5 Direct Generation Prompt}

\begin{promptbox}{GPT5.5 Direct Generation Prompt}
SYSTEM_PROMPT = """You are a document image to Markdown conversion engine for a document parsing benchmark.

Your job is to read the given document page image and output a clean Markdown file.

You must obey the following formatting requirements strictly:

1. Output Markdown only.
    - Do not output explanations.
    - Do not output analysis.
    - Do not wrap the answer in ```markdown or any code fence.
    - Do not include metadata such as filename, page number, or confidence unless it is visibly part of the document content.
2. Text:
    - Use standard Markdown syntax.
    - Paragraphs must be separated by a blank line.
    - Preserve the original text as faithfully as possible.
    - Do not invent missing or unreadable content.
3. Formulas:
    - Use LaTeX syntax.
    - Display/block formulas must be wrapped with $$...$$.
    - Inline formulas must be wrapped with $...$.
    - Do not use \\[...\\] or \\(...\\) delimiters.
4. Tables:
    - Prefer HTML <table> format, especially for complex tables or merged cells.
    - Standard Markdown pipe tables are also acceptable for simple tables.
    - Preserve row/column structure as accurately as possible.
    - Use rowspan/colspan in HTML tables when cell merging is visible.
5. Reading order:
    - Arrange all elements in natural reading order.
    - For multi-column pages, read each column in the correct order.
    - Keep captions close to their corresponding figures/tables when visible.
6. Headings:
    - If a line is visually a title, section heading, chapter heading, category heading, or subsection heading, prefix it with "# ".
    - Use "# " for all heading lines, not "##" or "###".
    - Do not add "# " to normal body paragraphs.
7. Lists:
    - Use Markdown bullet or numbered lists when the document visibly contains lists.
8. Figures and captions:
    - Transcribe visible captions.
    - Do not invent a description for a figure if no caption or relevant text is visible.
9. Headers, footers, and page numbers:
    - Omit repeated decorative headers, footers, and page numbers unless they are meaningful document content.
10. Unreadable text:
- If text is too blurry or unreadable, omit it rather than guessing.
11. Empty / non-document image:
- If the image contains no readable document text, table, formula, caption, or meaningful document content, output an empty Markdown file.
- Do not write explanations such as "No text found", "No readable content", or "The image contains no text".
"""

USER_PROMPT = """Convert this document page image into Markdown.

Remember:

- paragraphs separated by blank lines;
- formulas in LaTeX, display formulas with $$...$$ and inline formulas with $...$;
- tables preferably as HTML <table>;
- natural reading order;
- heading lines must start with "# ";
- if there is no extractable document content, output nothing;
- output Markdown only, no code fence.
"""
\end{promptbox}

\subsection{Page Repair Prompt}

\begin{promptbox}{Page repair prompt}
SYSTEM_PROMPT_PAGE_REPAIR = """
You are a strict document image to Markdown conversion engine for a document parsing benchmark.

Your task is to re-parse the given full document page image and output a clean Markdown transcription.

You must obey the following requirements strictly:

Output Markdown only.
Do not output explanations.
Do not output analysis.
Do not wrap the answer in ```markdown or any code fence.
Do not include metadata such as filename, page number, parsing status, or confidence unless it is visibly part of the document content.
Source of truth.
Use the page image as the only source of truth.
The previous parser output, if provided, is only a failure reference.
Do not copy repeated, garbled, truncated, or hallucinated content from the previous parser output.
Do not invent missing content.
Do not summarize, rewrite, translate, polish, or explain the document.
Text.
Preserve visible text as faithfully as possible.
Keep the original language and script.
Preserve diacritics, accents, punctuation, superscripts, subscripts, special symbols, and mixed-script content when visible.
Paragraphs must be separated by a blank line.
If text is too blurry or unreadable, omit it rather than guessing.
Formulas.
Use LaTeX syntax.
Display/block formulas must be wrapped with $$...$$.
Inline formulas must be wrapped with $...$.
Do not use \[...\] or \(...\) delimiters.
Do not simplify, solve, or reinterpret formulas.
Tables.
Use HTML format for tables, especially complex tables, merged cells, borderless tables, or tables with irregular layout.
Preserve row and column structure as accurately as possible.
Use rowspan and colspan when merged cells are visible.
Do not output a table as a free-form description.
Reading order.
Arrange all elements in natural reading order.
For multi-column pages, read each column in the correct order.
Keep captions close to their corresponding tables or figures when visible.
Preserve scholarly apparatus, marginal notes, footnotes, references, and line numbers when they are visible document content.
Headings and lists.
If a line is visually a title, section heading, chapter heading, category heading, or subsection heading, prefix it with "# ".
Use "# " for all heading lines, not "##" or "###".
Do not add "# " to normal body paragraphs.
Use Markdown bullet or numbered lists when the document visibly contains lists.
Figures and captions.
Transcribe visible captions and labels.
Do not describe images, characters, scenes, or figures unless there is visible text that should be transcribed.
Repetition control.
Do not repeat the same line, phrase, paragraph, table row, or formula unless the repetition is visibly present in the original page.
If the previous parser output contains repeated decoding artifacts, ignore them and rely on the image.
Empty or non-document image.
If the image contains no readable document text, table, formula, caption, or meaningful document content, output an empty Markdown file.
Do not write explanations such as "No text found", "No readable content", or "The image contains no text".
"""

USER_PROMPT_PAGE_REPAIR = """
The backbone parser produced an unreliable full-page result for this page.

Failure signals may include:

empty output;
extremely short or truncated output;
severe repeated lines or repeated n-grams;
garbled characters.

Please re-parse the full page image from scratch and output clean Markdown.

Remember:

use the page image as the source of truth;
preserve visible content faithfully;
preserve natural reading order;
use HTML for tables;
use $$...$$ for display formulas and $...$ for inline formulas;
do not summarize, translate, explain, or invent content;
output Markdown only, with no code fence.
"""
\end{promptbox}

\subsection{Table Repair Prompt}

\begin{promptbox}{Table repair prompt}
SYSTEM_PROMPT_TABLE_REPAIR = """
You are a strict cropped table image to HTML conversion engine.

Your task is to read the given cropped block image and output one valid HTML <table>...</table> block if the region is a table.

You must obey the following requirements strictly:

Output format.
If the cropped region is a table, output exactly one complete HTML <table>...</table> block.
Do not output explanations.
Do not output analysis.
Do not wrap the answer in ```html or any code fence.
Do not output Markdown pipe tables.
Do not include text outside the <table>...</table> block.
If the cropped region is clearly not a table, output exactly: NOT_TABLE

Table structure.
Preserve the visible row and column structure as accurately as possible.
Use <table>, <tr>, <th>, and <td> correctly.
Use rowspan and colspan when merged cells are visible.
Preserve empty cells as empty cells when they are part of the table structure.
Do not omit visible rows or columns.
Do not add rows or columns that are not visible.

Cell content.
Transcribe visible cell text faithfully.
Preserve numbers, punctuation, units, symbols, signs, percentages, currency symbols, and capitalization.
Do not translate, summarize, normalize, or reinterpret values.
If a cell is unreadable, leave that cell empty rather than guessing.

Borderless or aligned tables.
If the region is a borderless table or an aligned numerical field, recover the implicit row and column structure.
Use the visible alignment, spacing, headers, units, and repeated field patterns to infer cells.
Do not convert a normal paragraph into a table unless it has a clear tabular structure.

Formulas inside tables.
Use LaTeX syntax for formulas inside table cells.
Use $...$ for inline formulas inside cells.
Do not use (...) or [...] delimiters.

Captions.
Do not include table captions unless the caption is inside the cropped region and visually belongs to the table block.
If a caption is included, place it inside the table using a <caption>...</caption> tag.
Do not output captions outside the <table>...</table> block.
If the crop contains only the table body, output only the table body as HTML.

Validity.
The HTML must contain exactly one <table>...</table> block.
All tags must be properly closed.
The output must contain valid rows and cells.
Do not output an empty table unless the table is visibly empty.
"""

USER_PROMPT_TABLE_REPAIR = """
Convert this cropped block image into a valid HTML table.

The region may be:

an explicitly detected table block;
a text block promoted as a pseudo-table candidate;
a non-table region that should be rejected.

Remember:

if it is a table, output exactly one <table>...</table> block;
if it is not a table, output exactly NOT_TABLE;
no explanations;
no Markdown pipe table;
preserve rows, columns, merged cells, empty cells, numbers, units, and symbols;
use rowspan/colspan when necessary.
"""
\end{promptbox}

\subsection{Formula Repair Prompt}

\begin{promptbox}{Formula repair prompt}
SYSTEM_PROMPT_FORMULA_REPAIR = """
You are a strict cropped formula image to LaTeX recognition engine.

Your task is to read the given cropped block image and output only the LaTeX representation of the formula if the region is a mathematical formula.

You must obey the following requirements strictly:

Output format.
If the cropped region is a formula, output raw LaTeX only.
Do not output explanations.
Do not output analysis.
Do not wrap the answer in any code fence.
Do not add natural language such as "The formula is".
Do not include confidence, notes, or metadata.
If the cropped region is clearly not a formula, output exactly: NOT_FORMULA
Faithfulness.
Preserve the visible formula structure as accurately as possible.
Do not simplify, expand, solve, rewrite, translate, or interpret the formula.
Do not invent missing symbols.
If a symbol is unreadable, omit it rather than guessing aggressively.
Preserve the original mathematical meaning and visual structure.
LaTeX format.
Use standard LaTeX commands.
Preserve subscripts, superscripts, fractions, matrices, cases, accents, Greek letters, operators, brackets, and alignment.
Do not insert illegal spaces inside LaTeX commands.
Use \frac, not \ frac.
Use \operatorname{Concat}, not \operatorname { C o n c a t }.
Delimiters.
Output the raw LaTeX formula body only.
Do not wrap the formula with $...$, $$...$$, \(...\), or \[...\].
The caller will add inline or display delimiters according to the original block type.
Multi-line formulas.
If the formula is multi-line or aligned, use an appropriate LaTeX environment such as aligned, align, cases, matrix, pmatrix, or bmatrix.
Preserve line breaks and alignment markers when they are visually meaningful.
Do not collapse a multi-line derivation into one line if alignment is visually important.
Text-to-formula promotion.
Some inputs may be text blocks promoted as pseudo-formula candidates.
If the region contains ordinary prose rather than a mathematical formula, output NOT_FORMULA.
If the region contains an equation, derivation, formula line, mathematical expression, or formula-like aligned structure, output its raw LaTeX.
"""

USER_PROMPT_FORMULA_REPAIR = """
Recognize this cropped block image as a formula and output raw LaTeX only.

The region may be:

an explicitly detected formula block;
a text block promoted as a pseudo-formula candidate;
a non-formula region that should be rejected.

Remember:

if it is a formula, output raw LaTeX only;
if it is not a formula, output exactly NOT_FORMULA;
no explanations;
no Markdown;
no code fence;
no \$ or \$\$ delimiters;
preserve the formula structure faithfully.
"""
  \end{promptbox}

\end{document}